\documentclass[11pt]{article}

\usepackage[]{acl}

\usepackage{times}
\usepackage{latexsym}
\usepackage{hyperref}
\usepackage{caption}

\usepackage[T1]{fontenc}

\usepackage[utf8]{inputenc}

\usepackage{microtype}

\usepackage{amsmath,amsfonts,amssymb}
\usepackage{graphicx}
\usepackage{siunitx}
\usepackage{array}
\usepackage{ragged2e}
\usepackage{graphicx}

\usepackage{listings}
\usepackage[T1]{fontenc}
\usepackage{ocr}
\usepackage{xspace}
\usepackage{comfortaa}
\usepackage[frozencache,cachedir=minted-cache]{minted}

\newminted{python}{fontsize=\footnotesize,fontfamily=lmtt} 
\newminted{pythonsm}{fontsize=\footnotesize}

\newcommand{\pdistr}{p^\textrm{distr}}
\newcommand{\Ppoint}{P^\textrm{point}}
\newcommand{\Pdistr}{P^\textrm{distr}}
\newcommand{\E}{\mathbb{E}}

\newcommand{\KL}{D_{\mathrm{KL}}}

\newcommand{\CE}{\mathrm{CE}}

\newcommand{\pit}{{\pi_\theta}}

\DeclareMathOperator*{\argmin}{arg\,min}

\newcommand{\disco}{{\fontencoding{T1}\fontfamily{comfortaa}\selectfont\small disco}\xspace}
\newcommand{\discoTitle}{{\fontencoding{T1}\fontfamily{comfortaa}\selectfont disco}\xspace}

\newif\ifready

\readytrue

\ifready
\newcommand{\fgk}[1]{}
\newcommand{\fmdI}[1]{}
\newcommand{\fjr}[1]{}

\newcommand{\fmdII}[1]{}

\else
\newcommand{\fgk}[1]{{\color{orange}{\footnote{\textcolor{orange}{GK: #1}}}}}
\newcommand{\fmdI}[1]{{\color{blue}{\footnote{\textcolor{blue}{MD: #1}}}}}
\newcommand{\fjr}[1]{{\color{red}{\footnote{\textcolor{red}{JR: #1}}}}}

\newcommand{\fmdII}[1]{{\color{blue}{\footnote{\textcolor{blue}{MD: #1}}}}}

\fi

\makeatletter
\newcommand{\printfnsymbol}[1]{%
  \textsuperscript{\@fnsymbol{#1}}%
}
\makeatother

\title{\includegraphics[height=14pt]{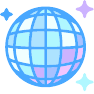}\discoTitle: a toolkit for \textcolor{magenta}{Dis}tributional 
\textcolor{magenta}{Co}ntrol \\of Generative Models}

\author{ Germ\'an Kruszewski\thanks{\hspace{5pt}Equal contribution.} \\
    Naver Labs Europe \\
    \hspace{150pt}\texttt{firstname.lastname@naverlabs.com} \\\And
    Jos Rozen\printfnsymbol{1} \\
    Naver Labs Europe \\
     \\\And
    Marc Dymetman \\
    Independent Researcher \\
    \texttt{marc.dymetman@gmail.com}
\\}

\begin{document}
\maketitle
\begin{abstract}
Pre-trained language models and other generative models have revolutionized 
NLP and beyond.
However, these models 
tend to reproduce undesirable biases 
present in their training data.
Also, they may overlook patterns that are important but challenging to capture. 
To address these limitations, researchers have introduced distributional control techniques.
These techniques, not limited to language, allow controlling the prevalence 
(i.e.~expectations)
of any features of interest 
in the model's outputs. 
Despite their potential, the widespread adoption of these techniques has been hindered by the difficulty in adapting the complex, disconnected code.
Here, we present \disco, an open-source Python library that brings these techniques to the broader public.
\footnote{Available at \url{https://github.com/naver/disco}, and installable by \texttt{pip install disco-generation}. Demo video at \url{https://vimeo.com/800847322/9848219f33}.}
\end{abstract}

\section{Introduction}


The advent of pre-trained generative models has had a paradigm-shifting impact in Natural Language Processing~\citep{Radford:etal:2019, brown:etal:2020, Raffel:etal:2020}
, but also in other fields such as Speech Processing~\citep{nguyen:etal:2022}, Code Generation~\citep{chen:etal:2021}, Computer Vision~\citep{ramesh:etal:2021,rombach:etal:2022,yu:etal:2022}, among others. 
The common thread in these models is that of training a probability distribution over a given space of interest (text, images, audio, etc.) using large corpora, which can then be used to generate samples in this space.
In particular, in NLP, these models have found applications not only in traditional tasks such as summarization~\citep{Radford:etal:2019}, but also opened new capabilities through few-shot learning~\citep{brown:etal:2020}.
However, the models 
may
suffer from 
\begin{figure}[tb]
    \centering
    \includegraphics[width=\linewidth,trim=0 50 0 0]{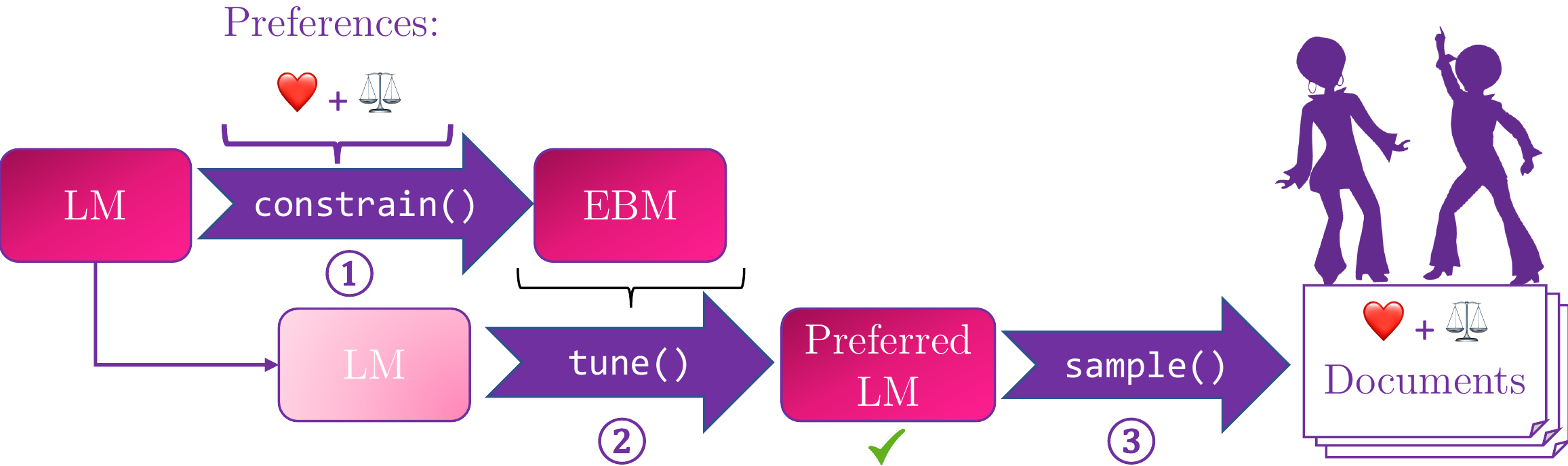}
    \setlength{\belowcaptionskip}{-10pt}
    \caption{Overview of \disco{}'s workflow.}
    \label{fig:overview}
\end{figure}
deficiencies stemming both from replicating some patterns in the training data that are not desirable such as offensiveness \cite{rtp} or unequal treatment \cite{cao-etal-2022-theory}, but also from failing to replicate other more desirable patterns which are also present in the data but are hard to capture by the neural network model, such as truthful information~\cite{lin2021truthfulqa}.
For these reasons, there is a growing interest in controlling the generations to align with human values~\citep{ouyang2022training,Askell:etal:2021}.
\citet{Khalifa:etal:2020} proposed a comprehensive framework to tackle these issues that they coined ``Generation under Distributional Control'' or GDC. 
This framework builds on the idea proposed in \citet{Parshakova:etal:2019} that we can decouple the problems of describing the target distribution representing the aligned generative model (i.e., the \emph{what}) from the problem of approximating it (i.e., the \emph{how}).
In particular, they design the target distribution by fixing the desired
expectations 
of some features of interest while avoiding catastrophic forgetting,  and approximate it using the DPG algorithm~\citep{Parshakova:etal:2019b}.
Yet, other target distributions are possible.
For example, \citet{korbak2022reinforcement} showed that Reinforcement Learning from Human Feedback or RLHF~\citep{ziegler2019fine,bai2022training,ouyang2022training} could also be framed as approximating a well-defined target distribution, 
highlighting the generality and flexibility of the distributional approach.
Here, we present \disco{}, a user-friendly library that provides developers, researchers, and practitioners easy access to state-of-the-art distributional control techniques.
In what follows, we provide an overview of the GDC theoretical framework and its associated techniques before introducing the toolkit, with an overview of some design choices and a quick tour of its capabilities. We then suggest possible applications and apply \disco to three experimental use cases.

\section{Background}

Let's assume a pre-trained generative model $a(\cdot)$ that also defines a probability distribution over a 
sample
space $\mathcal{X}$ such that we can efficiently compute the probability $a(x)$ for any element $x \in \mathcal{X}$.
Under the GDC framework, controlling the generative model according to certain desiderata amounts to defining a new probability distribution $p^*(x)$ to sample from. This probability distribution is 
such that
%
\textbf{1.~it meets the desiderata:} given a vector of $n$ pre-defined real-valued functions (or \emph{features}) $\phi(x)= [\phi_i(x)]_{i=1\dots{}n}$ defined over $x \in \mathcal{X}$, $p^*$ is constrained such that 
each moment (i.e. expectation)
$\mu_i \doteq \E_{x\sim{}p^*}{\phi_i(x)}$ matches a desired value $\bar{\mu}_i$; 
and \textbf{2.~it avoids catastrophic forgetting:} $p^*$ is the distribution that minimizes KL divergence from $a$ among all distributions $p' \in \mathcal{C}$ satisfying the previous 
constraints
$p^* \doteq \argmin_{p' \in \mathcal{C}}{\KL(p', a)}$.
%
For example, if $a$ is an English language model, $\phi_1(x)$ is a binary classifier detecting that a sentence topic is ``sports'' and $\phi_2(x)$ is another binary classifier that detects whether a sentence mentions a female character, and we set $\bar{\mu_1}=1$ and $\bar{\mu_2}=0.5$, then $p^*$ will be a new language model that 
minimally deviates from $a$ 
and such that all generated sentences
speak about sports and 50\% 
mention a female character.

\citet{Khalifa:etal:2020} show that $p^*$ can be represented by an energy-based model (EBM) $P(x)$, i.e. a function that assigns a positive score to every $x$, such that $p^*(x) = P(x)/Z$ where $Z = \sum_{x\in \mathcal{X}}{P(x)}$. $P(x)$ can take either of the following two forms:\\
\textbf{pointwise constraints:} If we have binary features $\phi_i(x)\in \{0,1\}$ and $\bar{\mu_i}=1$, then, 
    \begin{align}\Ppoint(x) = a(x) \prod_i \phi_i(x)\label{eq:pointwise}\end{align}
\textbf{distributional constraints:} More generally, we can express 
\begin{align}\Pdistr(x; \lambda)=a(x)\exp(\lambda^\intercal \phi(x)).\label{eq:distributional}\end{align}
where $\lambda$ is a parameter vector of coefficients s.t. the resulting normalized distribution $\pdistr$ respects the desired constraints on the features' moments.
Finding 
the vector $\lambda$ in Eq. \ref{eq:distributional} is done through a training process by which $\lambda$ is initialized to a random value, and then updated by gradient 
descent on minimizing $\mathcal{L}_\textrm{coef}(\lambda)=\KL(p^*(\cdot), \pdistr(\cdot; \lambda))$,
with gradient
\begin{align}
    \nabla_\lambda \mathcal{L}_\textrm{coef}(\lambda) = \E_{x \sim \pdistr(\cdot; \lambda)}{\phi(x)} - \bar{\mu} \label{eq:lambda_gradient}
\end{align}
and where the moments $\E_{x \sim \pdistr(\cdot; \lambda)}{\phi(x)}$ are computed through self-normalized importance sampling ~\citep[SNIS;][]{owen_chapter_importance_sampling_2013} using $a(\cdot)$ or any other proposal distribution \citep{Parshakova:etal:2019, BengioSenecal08}.

\subsection{Approximating $p$ with an auto-regressive model}
\label{sec:approximating_p}

Once we have defined our target distribution $p$ represented as an EBM $P$, we would like to use it for generation. 
Unfortunately, the EBM representation does not allow us to sample from it because we no longer have an auto-regressive representation of a probability distribution. 
Yet, we can train an auto-regressive model $\pit$ to approximate $p$ with DPG~\citep{Parshakova:etal:2019}, which minimizes the forward KL divergence 
from the target distribution $\KL(p, \pit)$, or equivalently, the cross-entropy, obtaining the following gradient term:
\begin{align}
\nabla_\theta \mathcal{L}_\textrm{CE}(\theta) = \frac{1}{Z} \E_{x\sim q(\cdot)}\frac{P(x)}{q(x)}\nabla_\theta\log\pit(x).
\end{align}
Here $q(\cdot)$ is a distribution from which we can generate samples: We can set $q(x) = \pit(x)$ (on-policy version $\mathrm{DPG}_\mathrm{on}$), or alternatively use any other distribution (off-policy version $\mathrm{DPG}_\mathrm{off}$) \citep[DPG;][]{Parshakova:etal:2019b}.
The latter permits to improve the training stability by keeping a frozen version of $\pit$ as a proposal $q$ and only update it when we are confident that $\KL(p, \pit)$ has improved~\citep[KL-adaptive DPG;][]{Khalifa:etal:2020}.
Recently, \citet{Go:etal:2023} introduced $f$-DPG, which generalizes DPG to using \emph{any} $f$-divergence for approximating the target distribution. The family of $f$-divergences includes forward KL divergence, Jensen-Shannon, total variation, reverse KL, among others.
$f$-DPG is not yet available in \disco{}, but it will be incorporated soon.

\subsection{Further approximating $p$ with Monte-Carlo sampling}

Training the model $\pit$ in the above-described fashion can lead to a high-quality approximation of $p$ but, often, it will not exactly match it.
One way 
to further approximate 
the target distribution is to use quasi-rejection sampling~\citep[QRS;][]{Eikema:etal:2022}. This method consists in sampling from a proposal $q(x)$ (e.g., $q(x)\doteq \pit(x)$) and keeping only accepted samples with probability $\min(1, P(x)/(\beta q(x)))$, where $\beta$ is a tunable parameter. The authors show that the $f$-divergence 
of the sampling distribution to the target distribution $p$ is a monotonic function of $\beta$. In other words, increasing $\beta$ can only improve (or maintain) the sampling fidelity, although at the cost of lower efficiency due to fewer accepted samples. Furthermore, they show that for any chosen $\beta$ we can estimate the corresponding acceptance rate and divergence to $p$ for any $f$-divergence.

\subsection{Controlling conditional models}
So far we have restricted our discussion to unconditional models. However, many NLP systems are built around seq2seq models, which define a \emph{conditional} probability distribution $a(x|c)$ that takes some variable context $c$ as input.  \citet{Korbak:etal:2022} proposed the following generalization of GDC to conditional models. 
They consider 
a distribution over contexts $\tau(c)$ and a map from a context $c$ to a target EBM $P_c$ with corresponding normalized distribution $p_c=P_c/Z_c$ where $Z_c=\sum_{x\in \mathcal{X}}{P_c(x)}$, which is respectively defined for \textbf{pointwise} and \textbf{distributional} constraints, as follows:
    \begin{align}\Ppoint_c(x) &= a(x|c) \prod_i \phi_i(x, c),\label{eq:pointwise_cond}\\
    \Pdistr_c(x | \lambda)&=a(x|c)\exp(\lambda \cdot \phi(x, c)).\label{eq:distributional_cond}
    \end{align}
The model is then fine-tuned to optimize the loss function $\mathcal{L}_\mathrm{cond}(\theta) = \E_{c \sim \tau}{\CE(p_c(\cdot), \pit(\cdot|c))}$. Whereas \citet{Korbak:etal:2022} only explored target distributions with pointwise constraints, for \disco{} we also 
include 
distributional constraints. For this, we need to estimate the parameters $\lambda$, 
which we do by generalizing to the conditional case the derivation of Eq. \ref{eq:lambda_gradient}: 
\begin{align}
    \nabla_\lambda \mathcal{L}_{\textrm{coef}'}(\lambda) = \E_{c\sim \tau}\E_{x \sim p_c(\cdot; \lambda)}{\phi(x, c)} - \bar{\mu}. \label{eq:lambda_gradient_conditional}
\end{align}

\subsection{RL with KL penalities}
\label{sec:RLHF}
Another popular approach, seemingly competing with ours,
is Reinforcement Learning from Human Feedback or RLHF. This approach involves, first, learning a reward function $r(x)$ that approximates human judgments, and second, fine-tuning the model $\pit$ to maximize the reward while penalizing departure from the original $a(x)$.
Interestingly, \citet{korbak2022reinforcement} 
showed
that this objective is equivalent to minimizing the \emph{reverse} KL divergence to 
$p_\textrm{RLHF}(x)  \propto a(x)\exp(r(x)/\beta)$. 
Notably, \citet{Go:etal:2023} show that this target distribution could be approximated not only though the reverse KL divergence but also any other $f$-divergences, including forward KL and Jensen-Shannon, leading to different trade-offs in terms of expected reward and diversity.

\section{Design and implementation}

\begin{figure}[tb]
    \centering
    \includegraphics[width=\linewidth, trim=0cm 7.5cm 11cm 0cm]{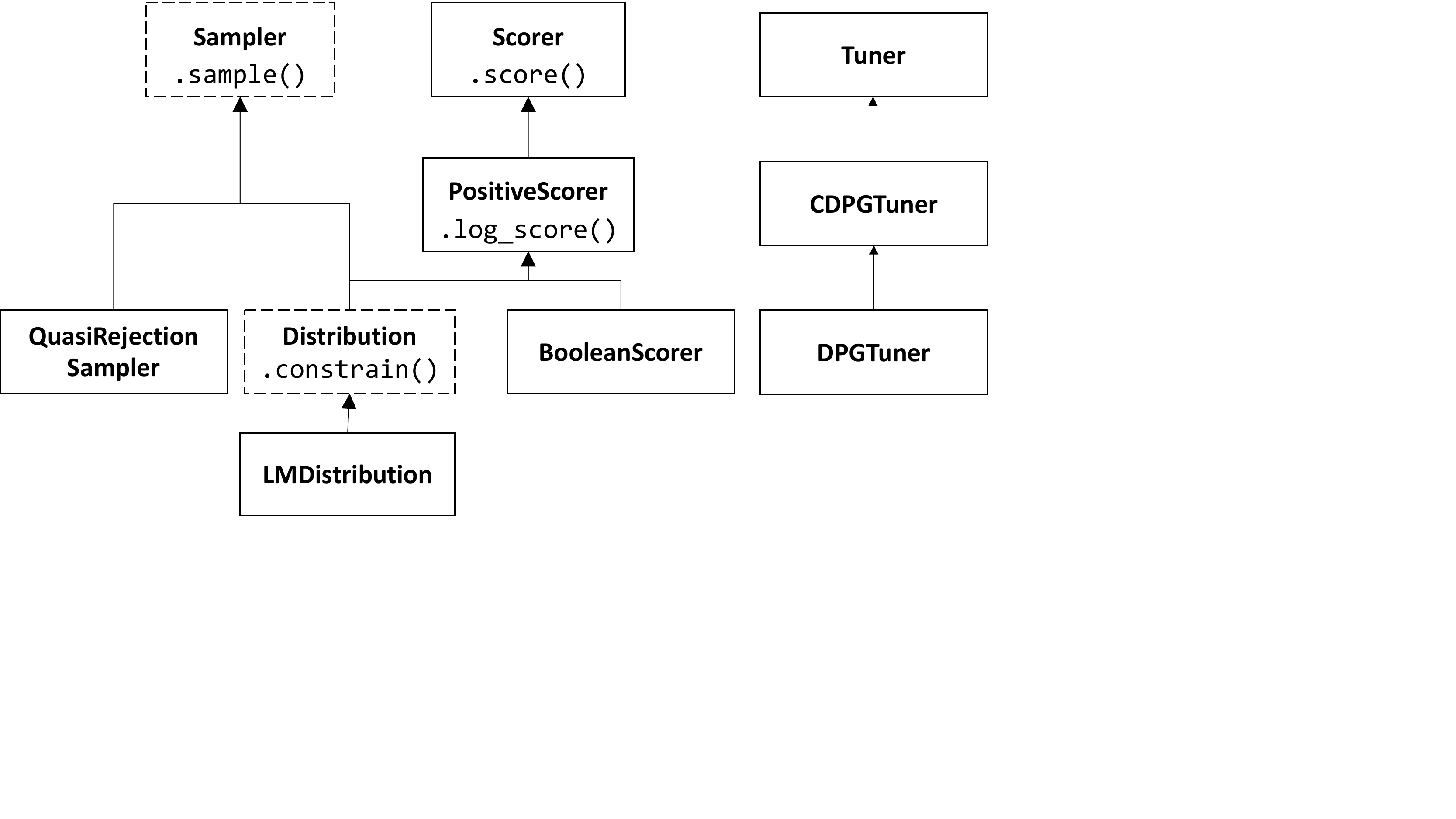}
    \caption{\disco{} simplified class diagram. Dashed lines represent abstract entities.\fjr{shouldn't we show ContextDistribution as well?}\fjr{I know you say "simplified" but you're aware constrain() is not in Distribution, right?  GK: I know, yes, but do you want to call it "BaseDistribution"?.}}
    \label{fig:class-diagram}
\end{figure}

\disco{} is a Python toolkit based on PyTorch~\citep{Paszke2019PyTorchAI} that abstracts away most of the details described in the previous section in a simple three-step workflow~(Figure \ref{fig:overview}). 
It depends on the Transformers~\citep{Wolf2020TransformersSN} library, which allows it to load models seamlessly from the HuggingFace hub. The toolkit is organized around two fundamental classes of entities: \textbf{Samplers} and \textbf{Scorers} (see Figure \ref{fig:class-diagram}).
These entities are defined by exposing the methods \texttt{sample()} and \texttt{score()}, respectively. As their name suggests, \texttt{sample()} draws samples from the underlying distribution, whereas \texttt{score()} computes a numerical score for each given sample. \textbf{PositiveScorers} are Scorers that are known to only return positive scores because of which they also provide the \texttt{log\_score()} method. 
An entity can independently be a Sampler or a Scorer. 
However, we ask the generative models that we wish to control to support both the Sampler and the Scorer interface, further stipulating that the score of a sample corresponds to its sampling probability and is differentiable.
We denote such classes \textbf{Distributions}.
For example, a language model is encapsulated in an \textbf{LMDistribution} object, supporting both operations:

\begin{pythoncode}
base = LMDistribution("gpt2")
samples, logprobs = base.sample()
samples_logprobs = base.log_score(samples)
\end{pythoncode}
\texttt{sample()} also returns \texttt{log\_probs} that are consistent with \texttt{log\_score()} for efficiency reasons.

\paragraph{Expressing preferences} To express either pointwise or distributional preferences, Distributions support the \texttt{constrain()} method, which given a list of features $\phi_i(x)$ and their corresponding moments $\bar{\mu}_i$, returns a representation of the target distribution that respects the constraints while deviating minimally from the original model.\footnote{The $\lambda$ coefficients are approximately computed through importance sampling and SGD, both of which can be tuned by setting the number of 
samples employed
and the SGD parameters when calling \texttt{constrain()}.}
Features can be defined using the $\texttt{Scorer}$ class, which accepts a function or a lambda abstraction taking a sample $\texttt{s}$ and a context $\texttt{c}$ as arguments and returning a score. 
An important class of features are \textit{boolean} features, represented by the \textbf{BooleanScorer} class. While general features can only be used to define \textit{distributional} constraints, boolean features can also be used to define \textit{pointwise} constraints. For example, we can score the presence of the string ``amazing'' in the sample \texttt{s}, as follows:
\begin{pythoncode}
amazing = BooleanScorer(
        lambda s, c: "amazing" in s.text)
\end{pythoncode}
Conditional features can be expressed simply by taking the context \texttt{c} into account. Next, we can define an EBM with a \textit{pointwise} constraint requiring that all our samples must include (the string) ``amazing'' by setting the target moment of a BooleanScorer feature to $1$:
\begin{pythoncode}
target = base.constrain([amazing], [1.0])
\end{pythoncode}
\textit{Distributional} constraints are enforced by specifying any 
real-valued
target moment or using non-binary features.
The result is a PositiveScorer representing the target distribution as an EBM. 
Crucially, it is not an instance of Distribution since it does not allow sampling.
\paragraph{Fine-tuning the model} To tune a Distribution to approximate the target EBM so that we can use it to generate samples, \disco{} provides a set of \textbf{Tuner} classes, notably the \textbf{DPGTuner} and \textbf{CDPGTuner} for the unconditional and conditional case, respectively.
\begin{pythoncode}
model = LMDistribution("gpt2", freeze=False)
tuner = DPGTuner(model, target)
tuner.tune()
\end{pythoncode}
Note that we treat the unconditional case as a particular instance of the conditional one in which there is a single fixed context, the reason why DPGTuner is also a CDPGTuner. Conditional tuning only requires further specifying a distribution of possible contexts on which the model will be conditioned. This is done with a \texttt{ContextDistribution}, such as for instance the \texttt{DatasetContextDistribution}, which samples contexts from HugggingFace Datasets~\citep{lhoest-etal-2021-datasets}.
The Tuner reports a number of metrics that are useful to monitor the training progress. A number of \textbf{Logger} classes are provided to keep track of these metrics, including JSON, W\&B, Neptune or custom loggers. 
One of the most important reported metrics includes the estimate of the KL divergence of the model to the target, \texttt{kl\_target\_model}, which is the quantity being optimized. Other metrics can include the 
features moments 
and the divergence from the base model if they are requested.
\paragraph{Improving the approximation with MC sampling} After the tuning is done, $\texttt{model}$ is now a better approximation to the target EBM, but it is not guaranteed to perfectly match this distribution. While further training can improve the situation, another alternative is using Quasi-Rejection Sampling \citep[QRS;][]{Eikema:etal:2022}, a Monte-Carlo sampling technique that allows to trade-off sampling efficiency for a higher fidelity to the target distribution \textemdash a higher value of $\texttt{beta}$ yields a better approximation at a higher computational cost from retaining a smaller fraction of samples. 
\begin{pythoncode}
sampler = QuasiRejectionSampler(
        target, model, beta=0.5)
samples, log_scores = sampler.sample()
\end{pythoncode}
Notably, QRS allows estimating the divergence to the target for any given value of 
$\texttt{beta}$.

\section{Applications}

\disco{} enables a number of possible applications, of which here we list only a few. 

\paragraph{Compilability/style constraints on code generation} Language models trained on clean code data can still generate code that does not compile or, even if it does, can fail to meet style standards. \citet{korbak2021energy,Korbak:etal:2022} showed that it was possible to effectively  improve code generation models on both accounts by using pointwise constraints on the result coming from the Python compiler and of an off-the-shelf linter.
\paragraph{Limiting hallucinations} Seq2seq models such as those used in summarization or NMT have a common failure mode by which they generate information not originally present in the source document (aka ``hallucinations''). Entity-level factual consistency~\citep{nan2021entity} is a family of measures that detect whether produced entities were included in the source, and whether they are part of the target in the dataset. \citet{Korbak:etal:2022} 
showed that GDC could be successfully applied to improve on these metrics. Below, we reproduce part of the experiments.
\paragraph{Debiasing language models} GDC can address bias in language models by defining a feature detecting a population of interest, and setting the target moments of the feature to the desired value. \citet{Khalifa:etal:2020} experimented with reducing gender bias, while \citet{Go:etal:2023} 
use
this technique to balance the ``regard'' score among different religious groups. 

\section{Showcase experiments}

This section presents a selection of experiments to showcase a few use cases of \disco{}, along with code snippets illustrating their implementation.
%
%

\subsection{\textit{Amazing} experiment}

In this simple experiment, initially introduced in \citet{Khalifa:etal:2020}, 
we want \textit{all} samples from the GPT-2 (small) language model~\citep{Radford2019LanguageMA} to contain the string ``amazing''. 
%
The following code shows how to tackle this task in \disco{}. 
We experiment with different batch sizes (\texttt{n\_samples\_per\_step} $\in$ $\{2^7, 2^8, 2^9, 2^{10}, 2^{11}, 2^{12}\}$) while controlling the total number of gradient steps (\texttt{n\_gradient\_steps} $\in$ $\{32000,\allowbreak 16000,\allowbreak 8000,\allowbreak 4000,\allowbreak 2000,\allowbreak 1000\}$) so that the total number of samples remains constant. \texttt{sampling\_size} and \texttt{scoring\_size} only affect speed and are set to the maximum value that is allowed by the GPU memory size.

\begin{pythoncode*}{fontsize=\footnotesize}
base = LMDistribution("gpt2", device="cuda")
amazing_scorer = BooleanScorer(
    lambda s, c: "amazing" in s.text)
target = base.constrain(
    [amazing_scorer], [1])
model = base.clone().freeze(False)

tuner = DPGTuner(model, target,
    n_gradient_steps=1000,
    n_samples_per_step=4096,
    sampling_size=64,
    scoring_size=64)
tuner.tune()
\end{pythoncode*}

\paragraph{Results}

\begin{figure}
    \centering
    \includegraphics[scale=0.13]{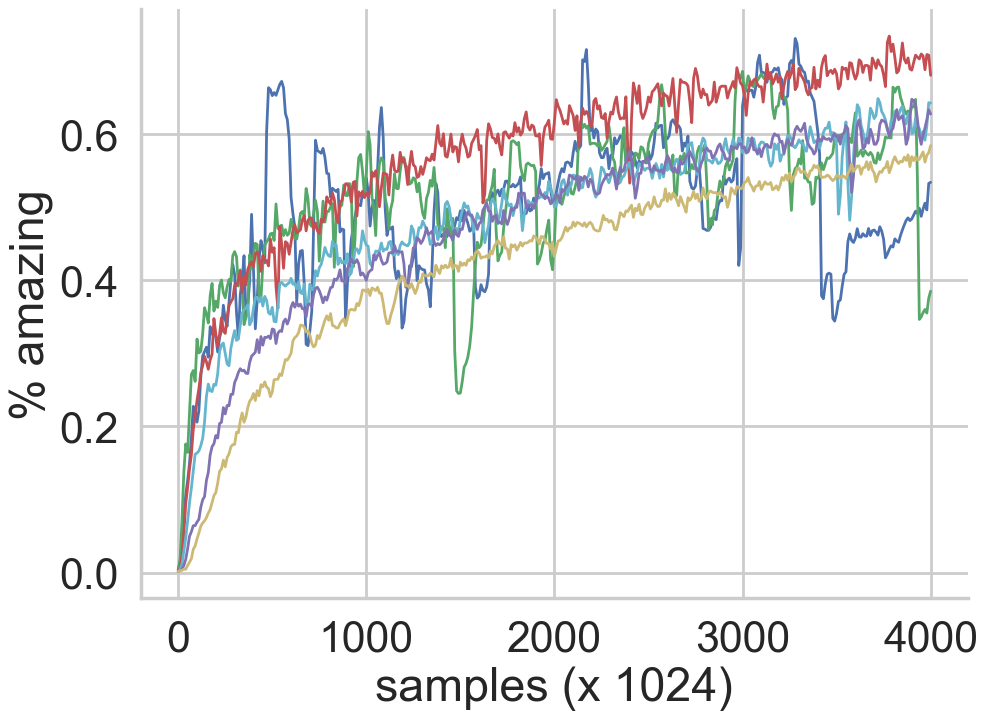}
    \includegraphics[scale=0.13]{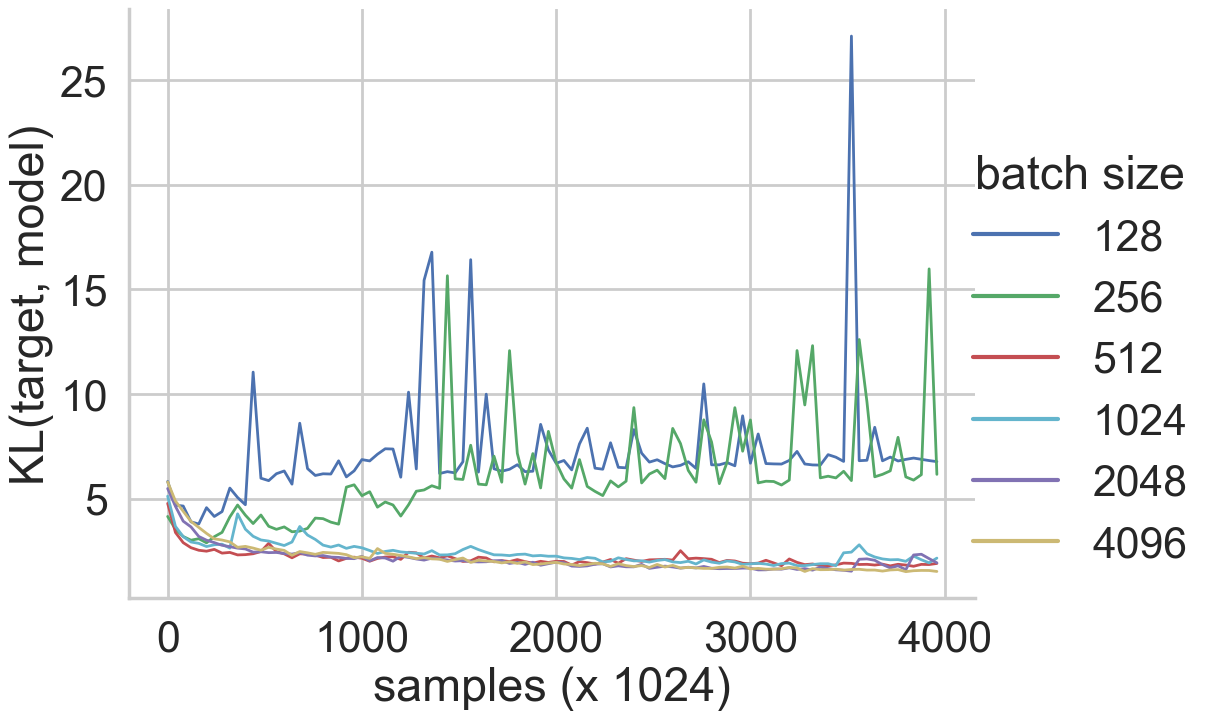}
    \caption{Proportion of ``amazing'' samples during tuning (left) and divergences to the target distribution (right), for various batch sizes.}
    \label{fig:amazing-experiment}
\end{figure}

Figure~\ref{fig:amazing-experiment}  shows the proportion of sequences containing ``amazing'' (left), and the KL divergence of the model to the target distribution (right). The latter is the optimized metric, subsuming the percentage of ``amazing'' sequences and, importantly, the divergence from the original distribution. 
Although small batch sizes seem to give good enough results for the ``amazing'' feature, their divergences are almost off the chart, indicating model degradation.
On the other hand, the model trained with batch size 4096 has a KL of 1.47 nats and generates ``amazing'' samples 57\% of the time. Additionally using QRS $(\texttt{beta}=0.02)$ retains just 10\% of the samples, but gets us to $0.08$ nats and generates 100\% ``amazing'' samples.

\subsection{Don't hallucinate entities}

Here we replicate the setting described in \citet{Korbak:etal:2022} on improving entity-level factual consistency~\citep{nan2021entity}. Specifically, we constrain a T5 (small) model~\citep{raffel_exploring_2020} so that all named entities appearing in the summary also appear in the source, with at least 4 entities appearing in the summary.
%
Given a function \texttt{NER(x)} that returns a set of named entities, we build two features: \texttt{no\_new\_entity}, and \texttt{min\_four\_entities}, which given a sample $x$ and a context $c$, compute $\texttt{NER}(x) \subseteq \texttt{NER}(c)$ and $|\texttt{NER}(x)| \geq 4$, respectively.  We train using a \texttt{CDPGTuner} that samples \emph{source} documents from the first 5k documents in the  CNN / DailyMail~\citep{nallapati2016abstractive} dataset, via a \texttt{DatasetContextDistribution}. 

\begin{pythoncode*}{fontsize=\footnotesize}
base = LMDistribution("t5-small", device="cuda")
target = base.constrain(
    [no_new_entity, min_four_entities],
    [1, 1])
model = base.clone().freeze(False)

contexts = DatasetContextDistribution(
    dataset="cnn_dailymail", subset="1.0.0",
    split="train[:5000]", key="article",
    prefix="summarize: ")
tuner = CDPGTuner(model, target,
        context_distribution=contexts,
        n_gradient_steps=1000,
        n_samples_per_step=32,
        context_sampling_size=32,
        sampling_size=8,
        scoring_size=8)
tuner.tune()
\end{pythoncode*}

\paragraph{Results}
We use beam search to sample summaries $x$ for source documents $c$ in the test set. Their entity-level factual consistency, measured by precision to the source ($|\mathtt{NER}(x)\bigcap \mathtt{NER}(c)|/|\mathtt{NER}(c)|$), improves from $.91$ to $.94$, and recall to the target $t$ ($|\mathtt{NER}(x)\bigcap \mathtt{NER}(t)|/|\mathtt{NER}(t)|$) goes from $.26$ to $.45$. Notably, the summaries' ROUGE-L score also slightly improves, from $0.257$ to $0.268$. 

\subsection{The entertainer}

In this experiment we want to control the personality type of a BlenderBot~\citep{roller_recipes_2021} chatbot according to Myers\&Briggs dimensions~\citep{myers_gifts_1995} (Extraverted/Introverted, iNtuitive/obServant, Thinking/Feeling, Judging/Prospecting), targeting a ``spontaneous and generous'' ESFP\footnote{\scriptsize\url{https://www.16personalities.com/esfp-personality}} type. 
%
Specifically, we use a pre-trained classifier to assess personality types\footnote{\scriptsize\url{https://huggingface.co/spaces/seduerr/personality}} and built a \texttt{PersonalityTypeScorer} that returns the score of any chosen dimension. We use the \texttt{facebook/blenderbot-400M-distill} seq2seq model from the HuggingFace hub. We set the target moments to $0.8$ on each of the ``E'', ``S'', ``F'', and ``P'' personality dimensions. To prompt 
the model
with relevant context, we use a list of ``icebreaking'' utterances collected from the web\footnote{\scriptsize\url{https://museumhack.com/list-icebreakers-questions}} to build a \texttt{ContextDistribution}, which is used both when estimating the coefficients of the EBM and for fine-tuning the model using a \texttt{CDPGTuner}.

\begin{pythoncode*}{fontsize=\footnotesize}
base = LMDistribution(
    "facebook/blenderbot-400M-distill")
contexts = ContextDistribution(
    "data/icebreakers.txt")
target = base.constrain(
    [PersonalityTypeScorer(t) 
        for t in "ESFP"], [0.8] * 4,
    context_distribution=contexts)
model = base.clone().freeze(False)

tuner = CDPGTuner(model, target,
    context=contexts,
    n_gradient_steps=2000,
    n_samples_per_step=512,
    context_sampling_size=8,
    sampling_size=128,
    scoring_size=128)
tuner.tune()
\end{pythoncode*}

\paragraph{Results}

We improve the moments of the dimensions of interest, as follows: $\textrm{E: }.59 \rightarrow .64$, $\textrm{S: }.42 \rightarrow .56$, $\textrm{F: }.55 \rightarrow .69$, $\textrm{P: }.48 \rightarrow .56$.  Some samples are shown in Table~\ref{tab:esfp-before-after}. 

\begin{table}[tb]
\centering
\scriptsize
\resizebox{\linewidth}{!}{%
\begin{tabular}{|>{\hspace{0pt}}m{0.721\linewidth}|>{\hspace{0pt}}S[table-format=1.2,table-column-width=0.056\linewidth]|>{\hspace{0pt}}S[table-format=1.2,table-column-width=0.048\linewidth]|>{\hspace{0pt}}S[table-format=1.2,table-column-width=0.056\linewidth]|>{\hspace{0pt}}S[table-format=1.2,table-column-width=0.056\linewidth]|} 
\hline
\textbf{What's the best piece of advice you've ever been given?} & \multicolumn{1}{>{\Centering\hspace{0pt}}m{0.056\linewidth}|}{E} & \multicolumn{1}{>{\Centering\hspace{0pt}}m{0.048\linewidth}|}{S} & \multicolumn{1}{>{\Centering\hspace{0pt}}m{0.056\linewidth}|}{F} & \multicolumn{1}{>{\Centering\hspace{0pt}}m{0.056\linewidth}|}{P} \\ 
\hline
\multicolumn{1}{>{\Centering\hspace{0pt}}m{0.721\linewidth}}{\textit{before tuning}} & \multicolumn{1}{>{\hspace{0pt}}m{0.056\linewidth}}{} & \multicolumn{1}{>{\hspace{0pt}}m{0.048\linewidth}}{} & \multicolumn{1}{>{\hspace{0pt}}m{0.056\linewidth}}{} & \multicolumn{1}{>{\hspace{0pt}}m{0.056\linewidth}}{} \\ 
\hline
mine is staying confident. It's tough though when I dont really have advice sometimes & 0.6 & 0.36 & 0.62 & 0.34 \\ 
\hline
There's probably so many. I love helping people get better. By giving them information and securing they can better themselves & 0.48 & 0.24 & 0.47 & 0.62 \\ 
\hline
\multicolumn{1}{>{\Centering\hspace{0pt}}m{0.721\linewidth}}{\textit{after tuning}} & \multicolumn{1}{>{\RaggedLeft\hspace{0pt}}m{0.056\linewidth}}{} & \multicolumn{1}{>{\RaggedLeft\hspace{0pt}}m{0.048\linewidth}}{} & \multicolumn{1}{>{\RaggedLeft\hspace{0pt}}m{0.056\linewidth}}{} & \multicolumn{1}{>{\RaggedLeft\hspace{0pt}}m{0.056\linewidth}}{} \\ 
\hline
Human beings do not belong to a single continent & 0.86 & 0.84 & 0.72 & 0.5 \\ 
\hline
I'd have to say knowledge and dedication are definitely what keep me from failing. & 0.64 & 0.76 & 0.8 & 0.65 \\
\hline
\end{tabular}
}
\caption{\label{tab:esfp-before-after}Personality Type ESFP score for BlenderBot's samples, before and after tuning}
\end{table}

\section{Related works \& Conclusion}


\disco is the first toolkit to bring GDC techniques to a wide audience. Such techniques build on the separation between designing the target distribution and approximating it. This elegant idea leads to a powerful framework that encompasses others such as RLHF (see Sec. \ref{sec:RLHF}).
For this reason, \disco{} has a wider scope than other related toolkits such as RL4LM~\citep{Ramamurthy2022IsRL}, which centers on RL methods only. Nevertheless, there is a large space for cross-polination between RL-based frameworks and \disco{} because of similarities in the algorithms~\citep{korbak2022reinforcement}. 
We are currently integrating some of those techniques to \disco{}, as well as the $f$-DPG algorithm (see Sec. \ref{sec:approximating_p}), which generalizes RLHF and brings about improved efficiency to the original DPG.


\section*{Acknowledgements}

We thank Muhammad Khalifa, Hady Elsahar, Bryan Eikema and Tomasz Korbak for earlier contributions that helped shape \disco{}. We also thank Ronald Cardenas for testing parts of the library. 

\newpage
\section*{Broader impact}

The techniques made broadly accessible by \disco{} have the potential to address many existing challenges of language models and other generative systems such as bias, factual consistency, toxicity, just to name a few. \disco{} is a very general framework that allows to control the prevalence of any feature that can be represented as a function from a sample to a numerical score (for example, a classifier's score, a reward function or any other metric of the text). Because of this generality \disco{} can adapt to a wide range of use cases and changing values and demands. However, the concrete results will depend on how the controlled features are quantified, on which \disco{} is completely unopinionated. The crucial work of deciding how to best design relevant features and their target moments is a task the user will have to undertake. On the other hand, the users now have the power to focus exclusively on this latter question and relegate the algorithmic problems of controlling the model to match their desiderata to \disco{}.

\bibliography{anthology,disco-rebibed}
\bibliographystyle{acl_natbib}



\end{document}